\documentclass[]{jingdong}

\usepackage{booktabs}
\usepackage[table]{xcolor}
\usepackage{makecell}
\usepackage{amsfonts}
\usepackage{nicefrac}
\usepackage{enumitem}
\usepackage{pifont}

\usepackage{amsmath,amssymb}
\usepackage{amsthm}
\usepackage{algorithm}
\usepackage{algorithmic}
\usepackage{colortbl}
\usepackage{wrapfig}
\usepackage[utf8]{inputenc}
\microtypesetup{expansion=false}
\graphicspath{{figures/}}
\title{PBSD: Privileged Bayesian Self-Distillation for Long-Horizon Credit Assignment}

\author[1,2,*]{Yang Tian}
\author[2]{Rui Wang}
\author[2]{Xumeng Wen}
\author[2]{Junjie Li}
\author[2]{Shizhao Sun}
\author[2]{Lei Song}
\author[2]{Jiang Bian}
\author[1,\dagger]{Bo Zhao}

\affiliation[1]{School of AI, Shanghai Jiao Tong University}
\affiliation[2]{XYZ AI Lab}

\contribution[*]{Work done during internship at XYZ.}
\contribution[\dagger]{Corresponding author.}
\checkdata[Email]{\email{yangtian6781@sjtu.edu.cn}}

\abstract{Long-horizon agentic tasks pose a fundamental credit assignment challenge for Reinforcement Learning (RL): trajectory-level rewards verify final correctness but provide limited guidance on which intermediate steps contribute to the outcome. The difficulty is especially pronounced in multi-turn search agents, where successful trajectories may contain misleading actions and failed trajectories may contain valuable evidence-gathering steps. We propose \textbf{PBSD} (\textbf{P}rivileged \textbf{B}ayesian \textbf{S}elf-\textbf{D}istillation), a Bayes-calibrated self-distillation method for fine-grained credit assignment under sparse final rewards. PBSD measures trajectory quality through the posterior-to-prior probability ratio of the verified answer and applies Bayes' rule to convert this hard-to-estimate answer-side ratio into a tractable likelihood ratio between a standard student model and a privileged answer-conditioned teacher model. The autoregressive decomposition of this Bayesian evidence score yields turn-level signals that identify whether each intermediate turn supports or undermines the verified outcome. Consequently, PBSD provides a principled and elegant reweighting scheme that transforms sparse outcome supervision into Bayes-calibrated turn-level credit signals, while remaining compatible with standard policy optimization. Experiments demonstrate that PBSD consistently enhances performance across both in-domain and out-of-domain settings, and effectively transfers knowledge from short-context training to long-context inference, suggesting that its fine-grained credit assignment mechanism facilitates more effective policy learning and yields improved generalization.}

\begin{document}

\maketitle

\section{Introduction}
\label{sec:intro}

Reinforcement Learning with Verifiable Rewards (RLVR) has emerged as an effective paradigm for improving Large Language Models (LLM) on tasks whose final outcomes can be automatically verified~\cite{deepseekr1,shao2024deepseekmath, grpo}. However, extending RLVR from single-turn reasoning to long-horizon agent training remains fundamentally challenging, particularly in search-agent settings~\cite{team2025tongyi,team2026dr}. Realistic search tasks require agents to interact with external tools over hundreds of turns while maintaining contexts that span tens to hundreds of thousands of tokens. These trajectories involve heterogeneous operations, including planning, query formulation, retrieval, tool invocation, evidence aggregation, and final response synthesis. Although a terminal reward can verify the correctness of the final answer, it provides sparse supervision for determining which intermediate actions contributed to the outcome. As a result, RLVR in such settings suffers from a severe credit assignment problem: successful trajectories may contain spurious, redundant, or misleading actions, while failed trajectories may still include useful reasoning or evidence-gathering steps.

Existing approaches to dense supervision face important limitations in this setting. Rubric-based~\cite{chen2026opensearch,team2026kimi} process supervision relies on strong evaluators and carefully designed criteria, making it costly and potentially biased when applied to long, heterogeneous trajectories. Heuristic process rewards~\cite{team2026mind} are cheaper but brittle and vulnerable to reward hacking, as agents may optimize superficial signals rather than genuinely outcome-relevant behavior. Tree-search-based~\cite{ji2025tree,yang2025treerpo} methods estimate intermediate credit by expanding future continuations, but their cost becomes prohibitive for trajectories with many tool interactions and long contexts. Distillation-based methods provide another route to dense supervision, but introduce their own challenges. On-policy distillation (OPD)~\cite{lu2025onpolicy} guides learning with a teacher policy on states visited by the student. However, its effectiveness depends not only on the teacher's capability, but also on its alignment with the student's reasoning patterns, decision style, and trajectory distribution~\cite{li2026rethinking}. When the teacher solves the task through substantially different reasoning paths or action preferences, its targets may be correct in isolation yet poorly matched to the student's on-policy behavior. Self-distillation~\cite{zhao2026opsd} with privileged information, such as the environmental feedback, avoids relying on an external teacher, but directly imitating the resulting privileged policy can introduce information leakage: the student may learn shortcuts or artifacts that are unavailable at inference time~\cite{yang2026self}. These limitations suggest a more precise objective: to exploit privileged information as evidence for credit assignment, rather than as a policy to be directly copied.

We propose \textbf{PBSD} (\textbf{P}rivileged \textbf{B}ayesian \textbf{S}elf-\textbf{D}istillation), a Bayes-calibrated self-distillation method for fine-grained credit assignment in long-horizon agents. PBSD evaluates each intermediate action by how it changes the probability of the verified outcome relative to the history-conditioned prior, yielding a posterior-to-prior evidence ratio over the final answer. Directly estimating this answer-side ratio is intractable, as it requires marginalizing over all possible future continuations from a given prefix. PBSD addresses this challenge by invoking Bayes' rule to reformulate the evidence ratio as Bayesian evidence scor: the likelihood of the observed action under a privileged answer-conditioned model relative to its likelihood under the standard student policy. This reformulation converts privileged outcome information into calibrated turn-level credit signals along the observed trajectory. Consequently, PBSD identifies whether each action supports or undermines the verified outcome without enumerating future trajectories, relying on external evaluators, or training the student to imitate privileged actions.

PBSD uses the Bayesian evidence score to calibrate the trajectory-level advantage at the turn level, rather than introducing an independent loss function. Thus, the terminal verifiable reward remains the source of the global learning signal, while intermediate actions are differentially weighted according to their Bayesian evidence for the verified outcome. This preserves the stability and simplicity of outcome-based RLVR, and because the privileged answer is used only through detached likelihood-ratio weights, PBSD avoids direct imitation of answer-conditioned behavior.

We validate PBSD on a 30B Mixture-of-Experts (MoE) model in long-horizon search-agent settings. Experimental results demonstrate that PBSD attains superior performance with substantially fewer training steps than outcome-only RLVR, while also exhibiting strong cross-dataset generalization, highlighting its ability to learn transferable credit signals for long-horizon agentic behavior. Our contributions are summarized as follows:
 
\begin{itemize}[nosep]
    \item We formulate turn-level credit assignment in RLVR-based search agents as Bayesian evidence estimation, evaluating intermediate actions by their evidence for the verified final outcome.
    
    \item We propose PBSD, a Bayes-calibrated self-distillation method that transforms an intractable posterior-to-prior ratio into an estimable privileged action-likelihood ratio for turn-level advantage calibration.
    
    \item We validate PBSD on a 30B MoE model, showing superior performance over several representative RLVR algorithms and strong cross-dataset generalization in long-horizon search-agent tasks.
\end{itemize}

\section{Related Work}

\subsection{Long-Horizon Credit Assignment in Agentic RL}
Credit assignment remains a central challenge in agentic reinforcement learning, where agents interact with environments through extended sequences of reasoning steps, tool calls, observations, and decisions. Compared with single-turn reasoning tasks, long-horizon agentic tasks involve extended sequences of heterogeneous reasoning and tool-use operations, where evidence is accumulated gradually and the relevance of each action is often only determined in hindsight. Consequently, outcome rewards provide only coarse supervision and offer limited guidance for attributing credit to individual intermediate steps.  Prior work has sought to provide denser supervision through rubric-based rewards and LLM-as-a-judge feedback~\cite{zhang2025criticsearch,chen2026opensearch,li2026rubricem,mahmoud2026reward}. However, these approaches typically depend on carefully engineered criteria and strong external evaluators. A separate line of work designs heuristic rule-based rewards for intermediate behaviors, such as verifying retrieved entities, measuring evidence coverage, or assessing tool-use patterns~\cite{team2026mind,wang2025stepsearch,wei2025reinforcing}. Although these rewards are cheaper to obtain, they require sufficiently complete and accurate task evidence to be converted into reliable intermediate signals. When the heuristic only captures part of what makes a trajectory useful, agents may learn to satisfy the rule itself rather than perform genuinely outcome-relevant reasoning or evidence gathering. Other approaches, including tree-search-based evaluation, aim to estimate step-level values more explicitly~\citep{yang2025treerpo,ji2025tree}. While promising, such methods can be computationally expensive in long-context search-agent settings, where accurate per-turn credit estimation may require numerous additional rollouts or value evaluations. In contrast, our work derives a lightweight Bayesian calibration signal directly from sampled trajectories, enabling turn-level credit assignment without handcrafted process rewards, external evaluators, or costly search-based credit estimation.

\subsection{On-Policy Distillation}
OPD combines on-policy data collection with token-level supervision: the student samples trajectories from its own policy, while a teacher provides guidance on the states induced by these trajectories~\citep{agarwal2024gkd,fu2026revisitingonpolicydistillationempirical}. Its effectiveness, however, depends not only on the teacher's capability but also on the compatibility between the teacher's reasoning pattern and the student's exploration distribution. When the teacher follows substantially different problem-solving strategies or action preferences, its token-level targets may be correct in isolation but poorly aligned with the student's on-policy behavior. Recent on-policy self-distillation methods~\cite{sdpo,ye2026policy,zhao2026opsd,shenfeld2026selfdistillation,penaloza2026privileged,sang2026policy} replace the external teacher with the same model conditioned on privileged information, such as reference answers. Although this removes the need for a stronger teacher, directly imitating privileged trajectories may introduce information leakage and destabilize training.

\subsection{Search Agents}
Search agents extend language models with external information access, allowing them to iteratively issue queries, inspect retrieved evidence, and synthesize final answers~\cite{chen2026opensearch,team2026dr,team2026mirothinker,team2025mirothinker}. Search-R1~\cite{jin2025search} trains models with outcome-based RL to interleave reasoning with multi-turn search queries, demonstrating that search behavior can be acquired without large-scale supervised trajectories. DeepResearcher~\cite{zheng2025deepresearcher} further scales RL for deep research agents in real-world web environments, highlighting the importance of training agents under authentic, noisy, and dynamic search interactions. These works demonstrate the value of RL for search-augmented reasoning, but they still largely rely on final-answer rewards or task-level success signals. In long-context search settings, where trajectories may contain many search turns and tool observations, such sparse rewards provide limited guidance about which intermediate searches were useful.
\section{Methodology}
\subsection{Preliminaries}
\label{sec:prelim}

\noindent \textbf{GRPO.}
Given an input query $x$, GRPO samples a group of $G$ trajectories from the old policy:
\[
\tau_i \sim \pi_{\theta_{\mathrm{old}}}(\cdot \mid x),
\qquad i=1,\dots,G .
\]
Each trajectory $\tau_i$ receives an outcome-level reward $r_i$. GRPO computes a group-relative advantage:
\[
A_i =
\frac{
r_i - \mathrm{mean}(\{r_j\}_{j=1}^G)
}{
\mathrm{std}(\{r_j\}_{j=1}^G) + \epsilon
}.
\]
The standard clipped GRPO objective can be written as:
\[
J_{\mathrm{GRPO}}(\theta)
=
\mathbb{E}
\left[
\frac{1}{G}
\sum_{i=1}^{G}
\frac{1}{|\tau_i|}
\sum_{k=1}^{|\tau_i|}
\min
\left(
\rho_{i,k}(\theta) A_i,
\mathrm{clip}(\rho_{i,k}(\theta),1-\eta,1+\eta) A_i
\right)
-
\beta D_{\mathrm{KL}}(\pi_\theta \| \pi_{\mathrm{ref}})
\right],
\]
where the token-level importance ratio is defined as:
\[
\rho_{i,k}(\theta)
=
\frac{
\pi_\theta(z_{i,k}\mid z_{i,<k},x)
}{
\pi_{\theta_{\mathrm{old}}}(z_{i,k}\mid z_{i,<k},x)
}
\]
For clarity, we consider the on-policy setting and omit the KL regularization term. In a multi-turn agent, each trajectory can be decomposed into a sequence of assistant turns:
\[
\tau_i = (a_{i,1},a_{i,2},\dots,a_{i,T_i}),
\]
where $a_{i,t}$ denotes the $t$-th assistant turn and $h_{i,t}$ denotes the interaction history before that turn. Under this decomposition, and ignoring the clipping operation, the GRPO gradient can be approximated as:
\[
\nabla_\theta J_{\mathrm{GRPO}}
\approx
\mathbb{E}
\left[
\frac{1}{G}
\sum_{i=1}^{G}
A_i
\sum_{t=1}^{T_i}
\sum_{k=1}^{|a_{i,t}|}
\nabla_\theta
\log \pi_\theta(a_{i,t,k}\mid h_{i,t},a_{i,t,<k})
\right].
\]
This formulation exposes a coarse credit-assignment scheme in which all tokens across every assistant turn within a trajectory share an identical trajectory-level advantage $A_i$. While this assignment is straightforward, it proves inadequate in long-horizon search-agent settings, where the contributions of individual turns are highly heterogeneous. As a result, the final outcome reward provides only a weak and noisy signal for determining whether each intermediate turn contributes positively or negatively to the final answer.

\subsection{PBSD: Privileged Bayesian Self-Distillation}

\begin{figure*}[t]
  \centerline{\includegraphics[width=0.88\textwidth, height=0.35\textheight]{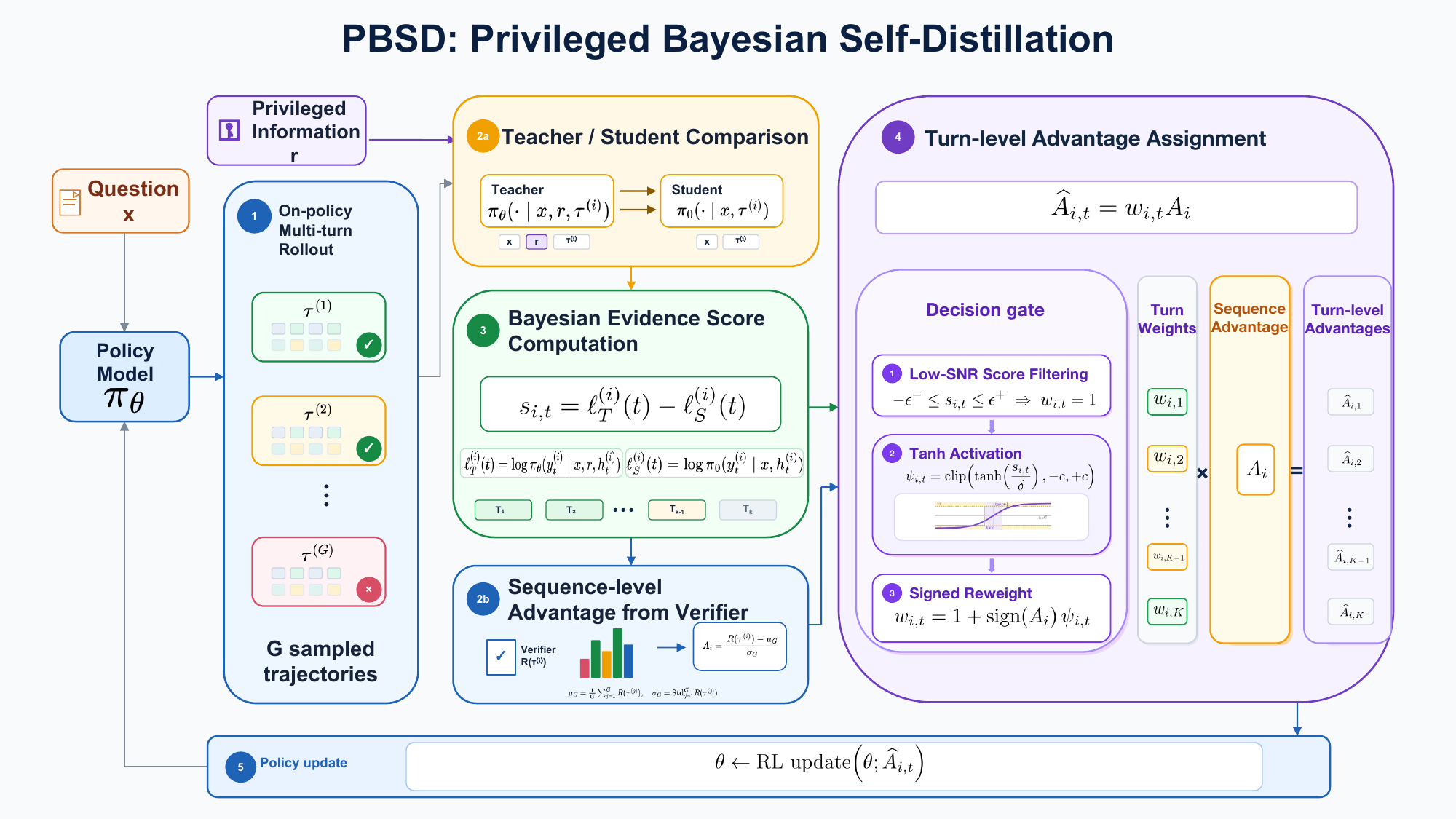}}
  \caption{Overview of Privileged Bayesian Self-Distillation. PBSD first performs on-policy multi-turn rollouts to obtain sampled reasoning trajectories, then compares teacher and student likelihoods using train-time privileged information to estimate turn-level Bayesian evidence score. The scores are combined with sequence-level advantages through a decision gate, producing calibrated turn-level advantages for policy update. }
  \label{main}
\end{figure*}

\noindent \textbf{From trajectory-level evaluation to turn-level evidence.}
Trajectory-level rewards indicate only whether a generated trajectory ultimately reaches the correct answer, offering a signal that is too coarse to discriminate among trajectories of differing quality or to localize which intermediate turns drive the final outcome. We first ask how its overall quality should be assessed before attributing credit to individual turns.

Intuitively, a desirable trajectory should make the verified answer \(y^\star\) more likely after being observed. That is, \(\tau_i\) provides positive support for \(y^\star\) if
\[
p(y^\star \mid x_i,\tau_i)
>
p(y^\star \mid x_i),
\]
and provides negative support if the inequality is reversed. Here,
\(p(y^\star \mid x_i)\) is the model's prior belief about the verified answer before observing the trajectory, while
\(p(y^\star \mid x_i,\tau_i)\) is the posterior belief after observing the trajectory. This motivates the trajectory support score
\[
S_i
=
\log
\frac{
p(y^\star \mid x_i,\tau_i)
}{
p(y^\star \mid x_i)
}.
\]
However, directly estimating the posterior probability \(p(y^\star \mid x_i,\tau_i)\) and the prior probability \(p(y^\star \mid x_i)\) is difficult. Because exact computation requires marginalizing over all possible latent paths, whose structures and probabilities are generally unobservable in practice. PBSD therefore applies Bayes' rule to rewrite the posterior-over-prior ratio as a likelihood ratio:
\[
\frac{
p(y^\star \mid x_i,\tau_i)
}{
p(y^\star \mid x_i)
}
=
\frac{
p(\tau_i \mid x_i,y^\star)
}{
p(\tau_i \mid x_i)
}.
\]
Thus, instead of estimating how likely the answer is before and after observing the trajectory, PBSD compares how likely the same trajectory is under two conditions: the standard student likelihood \(p(\tau_i \mid x_i)\), and the answer-conditioned teacher likelihood \(p(\tau_i \mid x_i,y^\star)\), where the model is given access to the ground-truth answer through a teacher-only conditioning prompt.

The trajectory-level log-likelihood ratio can then be decomposed into turn-level log-likelihood ratio Let
\(\tau_{i,<t}=(a_{i,1},\ldots,a_{i,t-1})\) denote the prefix before turn \(t\). By the autoregressive factorization of trajectory likelihoods,
\[
S_i
=
\sum_{t=1}^{T_i}
\left[
\log p(a_{i,t} \mid x_i,\tau_{i,<t},y^\star)
-
\log p(a_{i,t} \mid x_i,\tau_{i,<t})
\right].
\]

We define the turn-level Bayesian evidence score as
\[
s_{i,t}
=
\log p(a_{i,t}\mid x_i,\tau_{i,<t},y^\star)
-
\log p(a_{i,t}\mid x_i,\tau_{i,<t}).
\]
Thus, \(s_{i,t}\) measures the incremental contribution of turn \(a_{i,t}\) to the posterior evidence of the verified answer. In this way, PBSD converts trajectory-level correctness into fine-grained turn-level signal.


\noindent \textbf{Evidence-Calibrated Advantage Reweighting.}
PBSD does not directly use \(s_{i,t}\) as the optimization target. Instead, it uses \(s_{i,t}\) to calibrate how strongly each intermediate turn inherits this outcome-level signal:
\[
\widetilde A_{i,t}
=
w_{i,t} A_i,
\]
where the weight is given by a tanh-shaped modulation:
\[
w_{i,t} = 1 + \mathrm{sign}(A_i)\,
\mathrm{clip}\!\left(\tanh\!\left(\tfrac{s_{i,t}}{\delta}\right),\, -c,\, +c\right).
\]
Here, \(\delta>0\) controls the sensitivity of the modulation to the magnitude of the Bayesian evidence score, while the symmetric clipping threshold \(c\) constrains \(w_{i,t}\in[1-c,,1+c]\), thereby limiting the influence of outliers. This preserves the direction of the original GRPO signal: the final outcome still determines whether a trajectory is reinforced or penalized. The Bayesian evidence score only redistributes the magnitude across turns. In successful trajectories, evidence-supporting turns receive stronger reinforcement, while evidence-opposing turns are down-weighted. In failed trajectories, evidence-opposing turns receive stronger penalty, while evidence-supporting turns are protected from excessive punishment. PBSD therefore provides a principled log-likelihood-ratio signal for credit calibration.



\noindent \textbf{Low-SNR Score Filtering.}
Empirically, the per-turn evidence score \(s_{i,t}\) is concentrated around zero, where its sign is dominated by estimation noise rather than by genuine teacher--student disagreement. Propagating these low-magnitude scores into \(w_{i,t}\) injects a sign-aware perturbation that has no reliable directional meaning. We therefore introduce two filtering threshold and apply the modulation only when the evidence is sufficiently informative:
\[
w_{i,t}
=
\begin{cases}
1 + \mathrm{sign}(A_i)\,\mathrm{clip}\!\left(\tanh\!\left(\tfrac{s_{i,t}}{\delta}\right),\, -c,\, +c\right), & \ s_{i,t}<-\epsilon^{-}\ \text{or}\ s_{i,t}>\epsilon^{+} ,\\[2pt]
1, &   \quad\quad \epsilon^{-} \le \ s_{i,t}\ \le\epsilon^{+}.
\end{cases}
\]

Here, \(\epsilon^{+}>0\) and \(\epsilon^{-}>0\) are direction-specific thresholds for positive and negative \(s_{i,t}\), respectively, allowing the filter to account for asymmetric reliability across evidence directions.

\noindent \textbf{Non-Replay Evidence-Score Recomputation for R3-Routed MoE Models.}
For MoE models trained with R3 routing replay~\cite{ma2025stabilizing}, PBSD uses a separate likelihood-scoring pass for evidence-score recomputation. Although routing replay benefits policy optimization by reproducing the expert routes used during rollout, replayed logits are unsuitable for estimating the Bayesian evidence score. The evidence score should compare student and teacher likelihoods under the same non-replay evaluation procedure, so that any likelihood difference can be attributed to outcome conditioning rather than artifacts of routing replay. Therefore, when R3 replay is enabled, PBSD disables replay during evidence scoring: both student and teacher likelihoods are evaluated with fresh routing decisions, while the main policy update still uses standard replayed training log-probabilities.



Figure~\ref{main} illustrates the overall workflow of PBSD. PBSD provides a principled and elegant mechanism for converting outcome-level supervision into turn-level credit assignment without introducing an additional optimization objective. By using Bayesian evidence score only to calibrate the strength of inherited advantages, filtering low-SNR evidence near the decision boundary, and decoupling evidence scoring from MoE routing artifacts when necessary, PBSD preserves the stability of standard policy optimization while injecting more informative intermediate supervision. This design makes the method readily compatible with existing RL training pipelines and enables a controlled assessment of whether posterior evidence can improve the allocation of learning signal across multi-turn reasoning trajectories.

\section{Experiment}
\begin{table*}[t]
    \centering
    \begin{tabular}{lccccc}
        \toprule
        Method
            & Validation & BC (300) & BC (Easy) & BC (Medium)
            & BC (Hard) \\
        \midrule
        SFT Model
            & 31.75 & 29.83  
            & 67.75 & 20.25 & 1.50 \\
        GRPO
            & \underline{38.25} & \underline{32.33}
            & 71.00 & 23.75 & 2.25 \\
        OPSD
            & 33.25 & 30.17
            & \underline{70.00} & 18.00 & 2.50 \\
        GEAR
            & 36.50 & 32.00
            & 66.75 & \underline{26.25} & \underline{3.00} \\
        RLSD
            & 34.25 & 28.33
            & 64.25 & 19.00 & 1.75 \\
            \rowcolor[rgb]{0.87,0.94,1}
        PBSD
            & \textbf{40.87} & \textbf{35.83} & \textbf{74.50} & \textbf{28.50} & \textbf{4.50} \\
        \bottomrule
    \end{tabular}
\caption{Evaluation results on the in-domain validation set and the stratified subset of BrowseComp (BC) benchmark.}
\label{compare_algorithm}
\end{table*}

\begin{figure*}[t]
  \centerline{\includegraphics[width=0.93\textwidth, height=0.15\textheight]{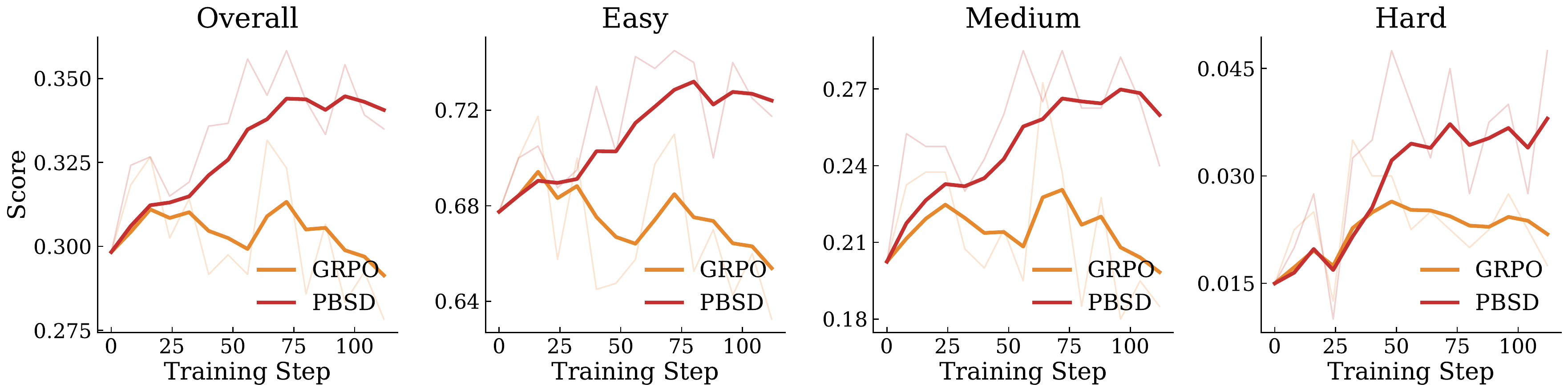}}
  \vspace{-0.1cm}
  \caption{Performance comparison of PBSD and GRPO on BrowseComp (300) across overall and difficulty-stratified splits. PBSD consistently achieves higher scores and stronger improvement trends across all settings.}
  \label{bc}
\end{figure*} 
\subsection{Experimental Setup}

\noindent \textbf{Training Data}
Our training corpus is constructed from two sources. First, we build a Wikipedia-derived knowledge graph and synthesize graph-grounded question-answering instances from it. We then employ MiroThinker-mini~\cite{team2026mirothinker} to generate search-agent trajectories for these instances, yielding approximately 2.1K trajectories. Second, we incorporate OpenSeeker~\cite{openseeker} as an additional data source and apply the same trajectory-generation procedure with MiroThinker-mini, collecting approximately 5.4K additional trajectories. In total, the resulting 7.5K trajectories are used for supervised fine-tuning (SFT).

For Reinforcement Learning (RL), we construct a separate synthetic dataset from the Wikipedia-derived knowledge graph, comprising 775 examples in total. We randomly reserve 200 examples as an in-domain validation set and use the remaining 575 examples for RL training. To ensure a rigorous and contamination-free evaluation protocol, the validation set is kept disjoint from the SFT corpus, and all RL examples are meticulously filtered and validated to eliminate overlap with the evaluation benchmark.

\noindent \textbf{Benchmark}
We evaluate PBSD on four challenging benchmarks that cover web browsing, multi-step information-seeking, and deep-search scenarios. \textbf{BrowseComp}~\citep{wei2025browsecomp} is a high-difficulty benchmark for browsing agents, where questions require persistent web navigation to locate hard-to-find and entangled information, while retaining short and easily verifiable answers. We use it as a primary benchmark for evaluating long-horizon information-seeking ability. \textbf{BrowseComp-ZH}~\citep{zhou2025browsecomp} extends this setting to the Chinese web, introducing multi-hop questions that require retrieval and reasoning over Chinese information sources. \textbf{GAIA}~\citep{mialon2023gaia} evaluates general AI assistants on real-world tasks involving reasoning, web browsing, and tool use; we use its text-only subset to focus on language-based search and reasoning. \textbf{xBench-DS-2505}~\citep{chen2025xbench} evaluates deep-search capabilities under the xBench framework, emphasizing planning, retrieval, reasoning, and answer synthesis. Together, these benchmarks provide a comprehensive evaluation of PBSD across long-horizon search, multilingual browsing, and generalizable agentic reasoning. For fair and robust evaluation, we report mean@4 on these benchmarks.

\noindent \textbf{Tool Server.}
During rollout, the agent interacts with an external tool server equipped with two tools: Serper for web search and Jina for webpage retrieval and content extraction. Serper is used to issue search queries and return ranked search results, while Jina fetches selected webpages, parses their content, and summarizes the extracted information using GPT-OSS-120B~\cite{agarwal2025gpt} as the summary model.

\noindent \textbf{Implementation Details.}
We use Qwen3-30B-A3B-Thinking-2507~\cite{yang2025qwen3} as the backbone model. During the SFT stage, we train the model using LlamaFactory~\cite{zheng2024llamafactory}. The learning rate is fixed at \(1\times10^{-5}\), the global batch size is 32. The model is fine-tuned for two epochs.

RL is performed within our in-house training framework, which uses Megatron~\cite{shoeybi2019megatron} as the training backend and SGLang~\cite{zheng2024sglang} as the rollout backend. The maximum context length is configured to 64K tokens. During rollout generation, each trajectory is allowed up to 300 interaction turns, with a per-turn generation budget of 16,384 tokens. For each prompt, we sample 8 trajectories and use a global batch size of 32 for training. Rollouts are generated with a temperature of \(1.0\) and top-\(p\) of 1.0. The policy is optimized with Adam, using an 8-step warm-up schedule and a peak learning rate of \(3\times10^{-6}\). The weight decay is set to \(0.1\). During evaluation, we set top-\(p\) to \(0.95\).

To ensure a fair comparison, we keep all non-algorithm-specific configurations identical across the compared methods, including the training data, rollout budget, maximum context length, sampling configuration, and learning-rate schedule. Algorithm-specific hyperparameters are set as follows. For GRPO-style objectives, we set the clipping thresholds to $\epsilon_{\mathrm{low}}=0.2$ and $\epsilon_{\mathrm{high}}=0.28$, without entropy regularization. For PBSD, the tanh modulation uses scale $\delta=0.1$ and clip $c=0.1$. The direction-specific deadband thresholds are set to \(\epsilon^{+}=0.001\) and \(\epsilon^{-}=0.003\)  The privileged teacher and the student are instantiated from the same base mode. PBSD only modifies the conditioning context used for likelihood evaluation and does not rely on an external teacher model. For OPSD, the teacher model is kept fixed throughout training, and the KL divergence is computed exactly over the full vocabulary rather than approximated from sampled tokens. For GEAR, we set the KL threshold to $0.4$ and the entropy threshold to $1.5$. For RLSD, we follow its default configuration: $\lambda$ is initialized to $0.5$ and linearly decayed to $0$ over the first 50 training steps, with $\epsilon_w=0.2$.

\subsection{Experimental Results}

\begin{table*}[t]
\centering
\small
\setlength{\tabcolsep}{6pt}
\renewcommand{\arraystretch}{1.15}
\begin{tabular}{lccccc}
\toprule
\textbf{Model} &
\makecell{\textbf{Train}\\\textbf{Data}} &
\makecell{\textbf{Browse}\\\textbf{Comp}} &
\makecell{\textbf{Browse}\\\textbf{Comp-ZH}} &
\makecell{\textbf{GAIA}\\\textbf{(Text-Only)}} &
\makecell{\textbf{xBench-}\\\textbf{DS-2505}} \\
\midrule

\multicolumn{6}{l}{\textbf{Foundation Models}} \\
\midrule
GLM-4.7          & -- & 67.5 & 66.6 & 61.9 & 72.0 \\
MiniMax-M2.1    & -- & 62.0 & 47.8 & 64.3 & 68.7 \\
DeepSeek-V3.2   & -- & 67.6 & 65.0 & 75.1 & \textbf{78.0} \\
Claude-4.5-Sonnet & -- & 24.1 & 42.2 & --   & --   \\
Claude-4.5-Opus & -- & \underline{67.8} & 62.4 & --   & --   \\
Seed-2.0-Pro & -- & \textbf{77.3} & \textbf{82.4} & \textbf{78.6}   & --   \\
OpenAI-o3       & -- & 49.7 & 58.1 & --   & 67.0 \\
GPT-5 High      & -- & 54.9 & 65.0 & \underline{76.4} & \underline{77.8} \\
Gemini-3-Pro    & -- & 37.8 & \underline{66.8} & --   & --   \\
\midrule
\multicolumn{6}{l}{\textbf{Trained Agents}} \\
\midrule
DeepDive-32B-SFT       & -- & 9.5  & 23.0 & --   & 48.5 \\
DeepDive-32B-RL        & -- & 14.8 & 25.6 & --   & 50.5 \\
MiroThinker-32B-v0.1-RL & 147k & 13.0 & 17.0  & -- & -- \\
WebSailor-V2-30B-RL    & -- & 35.3 & 44.1 & \underline{74.1} & 73.7 \\
WebLeaper-30B-RL       & -- & 38.8 & -- & -- & 72.0 \\
Tongyi-DR-30B          & -- & \textbf{43.4} & 46.7 & 70.9 & \textbf{75.0} \\
DeepMiner-32B-RL       & -- & 33.5 & 40.1 & 58.7 & 62.0 \\
OpenSeeker-30B         & 11.7k & 29.5 & \underline{48.4} & --   & \underline{74.0} \\
OpenResearcher-30B     & 97k & 26.3 & --   & 64.1 & 65.0 \\
REDSearcher-30B        & -- & 42.1 & \textbf{49.8} & \textbf{80.1} & --   \\
MindDR-v1.5-30B            & -- & \underline{42.8} & 45.7 & -- & \textbf{75.0}   \\
\midrule
\multicolumn{6}{l}{\textbf{Ours}} \\
\midrule
SFT & 7.5k & \underline{44.71} & 52.60 & 77.95 & 65.00 \\
SFT+GRPO   & 8k & 40.05 & \underline{55.36} & \underline{80.31} & \underline{67.00} \\
\rowcolor{blue!10}
\textbf{SFT+PBSD}& 8k & \textbf{46.21} & \textbf{57.44}\footnotemark & \textbf{81.10} & \textbf{71.00} \\
\bottomrule
\end{tabular}
\caption{Performance comparison across benchmarks. Within each model category, bold and underlined scores indicate the best and second-best performance on each benchmark}
\label{tab:benchmark-comparison}
\end{table*}

\footnotetext{We update the results after identifying outdated and erroneous annotations in the BrowseComp-ZH benchmark. The new results are obtained by evaluating on the corrected version of the dataset.}

To reduce evaluation costs while preserving a controlled assessment of cross-dataset generalization, we construct a stratified subset of BrowseComp, denoted BC(300). Concretely, we execute MiroThinker-mini five times on each BrowseComp example and partition the examples into three difficulty tiers based on empirical solve rates: examples answered correctly across all five runs are designated Easy, those answered incorrectly across all five runs are designated Hard, and the remaining examples are designated Medium. We then randomly sample 100 examples from each tier, yielding a balanced 300-example evaluation benchmark.

\textbf{Comparison with other methods.}
Table~\ref{compare_algorithm} reports the comparison between PBSD and several representative baselines, including the SFT model and RL algorithms. PBSD achieves the best results across all evaluation settings, with an accuracy of 40.87 on the in-domain validation set and 35.83 on BC(300). Compared with GRPO, PBSD yields absolute improvements of 2.62 and 3.50 points on the validation set and BC(300), respectively. The advantage of PBSD is also consistent across the stratified BrowseComp subsets. Notably, the gains are more pronounced on the easy and medium subsets, indicating improved robustness on challenging out-of-domain instances. As shown in Fig.~\ref{bc}, PBSD further exhibits more favorable optimization behavior than GRPO during the first 112 training steps, characterized by faster performance improvement, higher final accuracy, and more stable convergence across all BC subsets. These results demonstrate that PBSD not only enhances in-domain performance over standard RLVR methods, but also improves generalization to out-of-domain evaluation settings.

\begin{figure*}[t]
  \centerline{\includegraphics[width=0.95\textwidth, height=0.23\textheight]{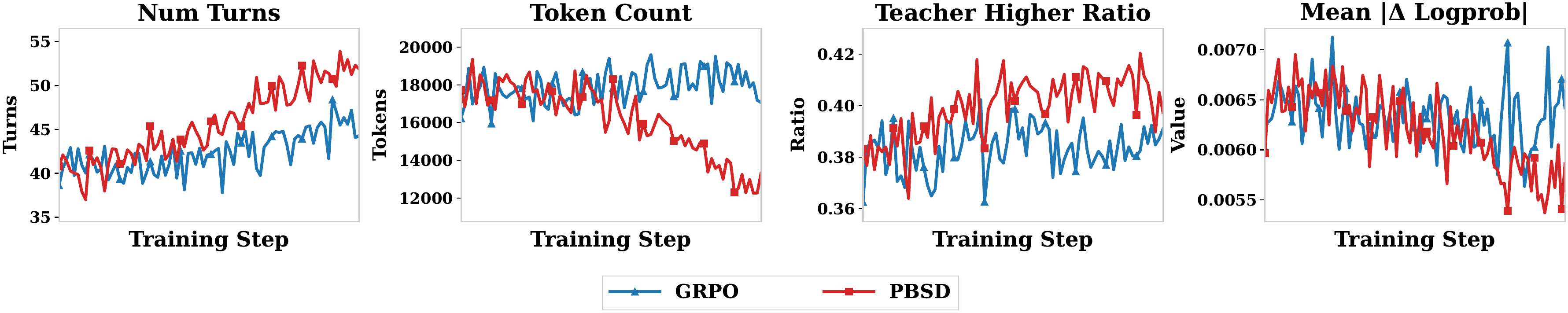}}
  \vspace{-0.1cm}
  \caption{Training dynamics of GRPO and PBSD.}
  \label{Phenomenon}
\end{figure*}

\textbf{Cross-Benchmark Generalization under Long-Context Evaluation.}
Table~\ref{tab:benchmark-comparison} compares PBSD with recent foundation models and search agents. Although trained under a 64K context window, PBSD is evaluated under a 256K context setting, testing its transferability beyond both the training distribution and training context length. This setting is particularly challenging on BrowseComp, where successful problem solving often requires longer-horizon exploration and more rounds of tool invocation. Under this demanding regime, SFT+PBSD achieves the best BrowseComp performance among trained agents and consistently improves over the SFT baseline across all benchmarks. In contrast, SFT+GRPO improves mainly on benchmarks involving fewer search turns, while its performance drops on BrowseComp. This suggests a key limitation of sparse-reward optimization in GRPO: when the search trajectory becomes longer and credit assignment more ambiguous, outcome-level rewards provide insufficient guidance for optimizing intermediate decisions. By introducing Bayesian turn-level credit calibration, PBSD provides more informative supervision for long-horizon search behavior, leading to stronger performance on BrowseComp and more robust generalization across BrowseComp-ZH, GAIA (Text-Only), and xBench-DS-2505.

\begin{wrapfigure}[16]{r}{0.5\linewidth}
  \centering
  \vspace{-20pt}
  \includegraphics[width=0.45\textwidth, height=0.25\textheight]{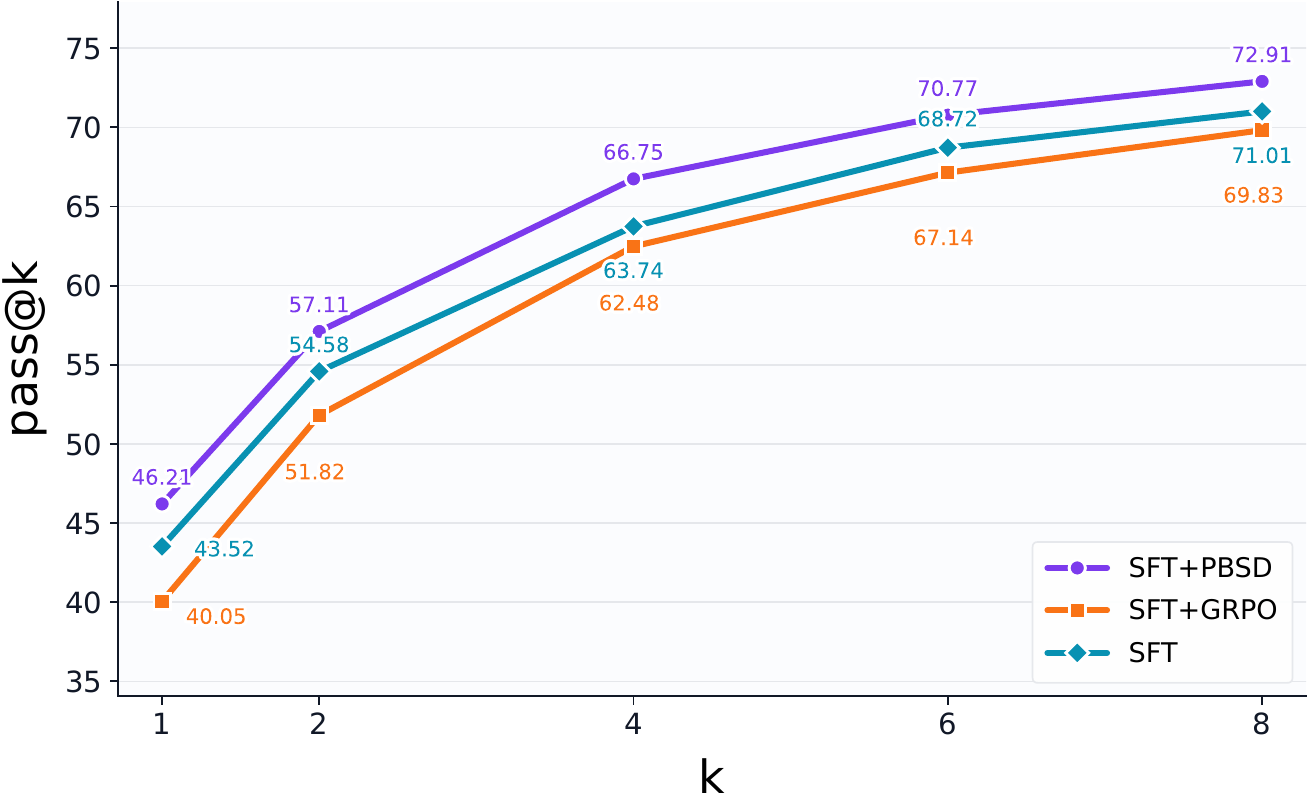}
  \caption{Pass@k comparison of SFT, SFT+GRPO, and the proposed SFT+PBSD method on BrowseComp.}
  \label{passk}
\end{wrapfigure} 

\textbf{Capability Boundary Analysis.}
Figure~\ref{passk} reports Pass@k under increasing sampling budgets. SFT+PBSD consistently outperforms the SFT baseline across all values of $k$, indicating that PBSD improves not only the most likely response but also the broader distribution of sampled trajectories. Since Pass@k reflects whether a correct solution is reachable within a finite sampling budget, the gains at larger $k$ suggest that PBSD increases the density of valid reasoning paths and expands the model's effective capability boundary. In contrast, SFT+GRPO remains below SFT throughout, implying that sparse outcome-reward optimization may over-concentrate the policy and suppress useful alternative solution modes. This contrast is especially relevant for long-context search tasks, where robust performance depends on maintaining diverse intermediate routes rather than committing prematurely to a narrow trajectory family.

\textbf{Behavioral Dynamics of PBSD during Training.}
To better understand the source of these gains, we analyze the behavioral dynamics induced by PBSD during training. As shown in Figure~\ref{Phenomenon}, PBSD exhibits a distinct pattern compared with GRPO: it increases the number of interaction turns while substantially reducing the total number of generated tokens. This suggests that PBSD shifts the agent's computation from verbose free-form generation toward more frequent, focused, and evidence-oriented search interactions, a desirable property for long-horizon search agents. In addition, the fraction of turns where the privileged answer-conditioned likelihood exceeds the ordinary student likelihood steadily increases over training, indicating that the student's sampled actions become increasingly consistent with outcome-relevant behavior. We further observe that the absolute teacher--student likelihood gap gradually decreases, suggesting that the teacher increasingly recognizes student-generated intermediate actions as supportive of the verified answer. Notably, this reduction in teacher--student discrepancy is achieved without directly optimizing a KL divergence or distilling the privileged answer-conditioned policy, thereby mitigating the risk of privileged-information leakage. Together, these results show that PBSD not only improves final-task performance, but also reshapes the agent's interaction strategy toward more efficient evidence gathering and stronger alignment with outcome-relevant behavior.

\subsection{Ablation Study}

\begin{table*}[t]
    \centering
    \resizebox{0.9\textwidth}{!}{
    \begin{tabular}{lccccc}
        \toprule
        Variant
            & Validation & BC (300) & BC (Easy) & BC (Medium)
            & BC (Hard) \\
        \midrule
        \rowcolor[rgb]{0.87,0.94,1}
        PBSD
            & \textbf{40.87} & \textbf{35.83}
            & \textbf{74.50} & \textbf{28.50} & {4.50} \\
        \midrule
        \multicolumn{6}{l}{\textit{(a) Non-Replay Evidence-Score Recomputation}} \\
        \quad w/o Non-Replay Evidence-Score Recomputation
            & {27.75} & {29.08} & {64.50} & {19.75} & {3.00} \\
        \midrule
        \multicolumn{6}{l}{\textit{(b) Soft-Modulation Scale $\delta$.}} \\
        \rowcolor[rgb]{0.87,0.94,1}
        \quad $\delta=0.1$ \
            & 40.87 & 35.83 & 74.50 & 28.50 & 4.50 \\
        \quad $\delta=0.5$
            & {37.25} & {35.33} & \textbf{75.50} & {25.50} & {5.00} \\
        \quad $\delta=1.0$
            & {37.25} & {33.67} & {70.50} & {27.5} & {3.00} \\
        \quad $\delta=2.0$
            & {37.13} & {32.00} & {73.00} & {20.50} & {2.50} \\
        \midrule
        \multicolumn{6}{l}{\textit{(c) Low-SNR Filtering Thresholds $(\epsilon^{+},\epsilon^{-})$}} \\
        \quad $(\epsilon^{+}=0,\epsilon^{-}=0)$ \quad (0\% filtered)
            & {34.87} & {33.25} & {72.00} & {24.50} & {3.25} \\
        \quad $(\epsilon^{+}=0.0005,\epsilon^{-}=0.001)$ \quad ($\sim$10\% filtered)
            & {35.25} & {33.58} & {72.25} & {25.75} & {2.75} \\
        \quad $(\epsilon^{+}=0.001,\epsilon^{-}=0.002)$ \quad ($\sim$20\% filtered)
            & {40.50} & {34.00} & {72.00} & {24.00} & \textbf{6.00} \\
        \rowcolor[rgb]{0.87,0.94,1}
        \quad $(\epsilon^{+}=0.001,\,\epsilon^{-}=0.003)$  \quad ($\sim$30\% filtered)
            & 40.87 & 35.83 & 74.50 & 28.50 & 4.50 \\
        \quad $(\epsilon^{+}=0.0005,\,\epsilon^{-}=0.007)$ \quad ($\sim$40\% filtered)
            & {36.75} & {33.00} & {70.5} & {24.25} & {4.25} \\
        \bottomrule
    \end{tabular}
    }
    \caption{Ablations on the three core design choices of PBSD.}
    \label{tab:ablation_pbsd}
\end{table*}

Table~\ref{tab:ablation_pbsd} ablates the three core design choices of PBSD on the in-domain validation set and the stratified BrowseComp (BC) subset.
\textbf{(a)~Non-Replay Evidence-Score Recomputation.}
On MoE backbones the teacher--student likelihood gap \(s_{i,t}\) is contaminated by stochastic routing noise unless evidence scoring is recomputed under fresh routing. Removing this recomputation collapses validation accuracy from \(40.87\) to \(27.75\) and drops BC(300) by \(6.75\) points, confirming that replay-free recomputation is essential rather than incidental for MoE-based evidence scoring.
\textbf{(b)~Soft-Modulation Scale \(\delta\).}
The scale \(\delta\) controls how sharply the tanh modulation responds to the magnitude of \(s_{i,t}\). Both extremes are suboptimal: a small \(\delta\) saturates the modulation on noisy turns, while a large \(\delta\) (\(\delta=2.0\)) flattens it into a near-uniform reweighting and recovers GRPO-like behaviour, yielding a clear regression on BC(Medium) and BC(Hard); \(\delta=0.1\) lies near the inflection point of the empirical \(|s_{i,t}|\) distribution and consistently performs best.
\textbf{(c)~Low-SNR Filtering Thresholds.}
The deadband thresholds \((\epsilon^{+},\epsilon^{-})\) gate out turns whose evidence is dominated by estimation noise. Without filtering (\(0\%\) filtered) the model gains a sign-aware perturbation with no reliable directional meaning and trails the full method by more than \(6\) points on validation accuracy. Filtering only the bottom \(10\%\) is insufficient, while filtering up to \(40\%\) discards informative turns and degrades all metrics. The asymmetric setting \((\epsilon^{+}=0.001,\,\epsilon^{-}=0.003)\), which removes roughly \(30\%\) of the lowest-SNR turns and reflects the natural asymmetry between teacher-higher and teacher-lower evidence, achieves the best overall trade-off.
Across the three axes, every component contributes a measurable gain over its ablated counterpart.

\section{Conclusion}

In this work, we studied the credit assignment problem in RLVR-based long-horizon search agents and introduced PBSD, a Bayes-calibrated self-distillation method that derives turn-level credit from privileged answer-conditioned likelihoods. By interpreting trajectory quality as a posterior-to-prior evidence ratio and applying Bayes' rule, PBSD converts the otherwise intractable answer-side estimation problem into an estimable action-likelihood ratio, which is then used as a detached calibration weight for trajectory-level advantages. Empirically, PBSD consistently outperforms outcome-only RLVR and strong baselines, while exhibiting robust cross-dataset and long-context generalization. These results highlight Bayesian evidence calibration as an effective mechanism for improving credit assignment in long-horizon agent training.

\section{Limitations and Future Work}

PBSD relies on access to verified final answers to construct privileged answer-conditioned likelihoods for Bayesian evidence estimation. This assumption is well aligned with RLVR settings such as search QA and verifiable reasoning tasks, where final outcomes can be automatically checked. However, in more open-ended agentic scenarios, ground-truth answers may be ambiguous, incomplete, or admit multiple valid solutions, making the privileged conditioning signal harder to define. Moreover, PBSD assumes that the model used to compute answer-conditioned likelihoods is sufficiently capable and well calibrated; otherwise, the resulting evidence ratios may reflect likelihood noise or model miscalibration rather than reliable turn-level credit. Future work could relax these requirements by replacing explicit ground-truth conditioning with learned verifiers, stronger teacher models, or other evaluative signals, enabling Bayesian credit calibration in broader agentic tasks without requiring direct access to unique verified answers.

\bibliographystyle{unsrtnat}
\bibliography{neurips_2026}

\newpage
\appendix

\section{Search Agent System Prompt}
\label{app:search-agent-system-prompt}

The search agent is initialized with the following system prompt. The prompt defines its tool-use protocol, available MCP-style tools, and high-level search objective

\definecolor{skyblue}{RGB}{91,155,213}

\begin{tcblisting}{
    enhanced,
    breakable,
    colback=lightgray!20,
    colframe=skyblue,
    title=Search Agent System Prompt,
    fonttitle=\bfseries,
    listing only,
    listing options={
        basicstyle=\ttfamily\scriptsize,
        breaklines=true,
        breakindent=0pt,
        breakautoindent=false,
        postbreak={},
        prebreak={},
        columns=fullflexible,
        keepspaces=true,
        showstringspaces=false
    }
}
In this environment you have access to a set of tools you can use to answer the user's question. \

You only have access to the tools provided below. You can only use one tool per message, and will receive the result of that tool in the user's next response. You use tools step-by-step to accomplish a given task, with each tool-use informed by the result of the previous tool-use. Today is: {formatted_date}

# Tool-Use Formatting Instructions

Tool-use is formatted using XML-style tags. The tool-use is enclosed in <use_mcp_tool></use_mcp_tool> and each parameter is similarly enclosed within its own set of tags.

The Model Context Protocol (MCP) connects to servers that provide additional tools and resources to extend your capabilities. You can use the server's tools via the `use_mcp_tool`.

Description:
Request to use a tool provided by a MCP server. Each MCP server can provide multiple tools with different capabilities. Tools have defined input schemas that specify required and optional parameters.

Parameters:
- server_name: (required) The name of the MCP server providing the tool
- tool_name: (required) The name of the tool to execute
- arguments: (required) A JSON object containing the tool's input parameters, following the tool's input schema, quotes within string must be properly escaped, ensure it's valid JSON

Usage:
<use_mcp_tool>
<server_name>server name here</server_name>
<tool_name>tool name here</tool_name>
<arguments>
{{
"param1": "value1",
"param2": "value2 \\"escaped string\\""
}}
</arguments>
</use_mcp_tool>

Important Notes:
- Tool-use must be placed **at the end** of your response, **top-level**, and not nested within other tags.
- Always adhere to this format for the tool use to ensure proper parsing and execution.

String and scalar parameters should be specified as is, while lists and objects should use JSON format. Note that spaces for string values are not stripped. The output is not expected to be valid XML and is parsed with regular expressions.
Here are the functions available in JSONSchema format:
{mcp_server_definitions}
# General Objective

You accomplish a given task iteratively, breaking it down into clear steps and working through them methodically.

# Agent Specific Objective

You are a task-solving agent that uses tools step-by-step to answer the user's question. Your goal is to provide complete, accurate and well-reasoned answers using additional tools.
\end{tcblisting}

\captionof{figure}{System prompt used by the search agent.}
\label{fig:search-agent-system-prompt}

\section{Pseudocode}

Algorithm~\ref{alg:pbsd} summarizes the training procedure of PBSD.
For each prompt, we first sample a group of trajectories and compute the standard group-normalized advantage from verifier rewards.
PBSD then constructs a privileged scoring context by augmenting the original prompt with the ground-truth answer, and evaluates the same generated tokens under both the original and privileged contexts.
The difference between these two likelihoods provides a turn-level self-distillation signal: turns that become more likely under the privileged context are softly amplified when their trajectory advantage is positive, and suppressed when their trajectory advantage is negative, while the opposite adjustment is applied when the privileged context assigns lower likelihood.
The privileged information is therefore used only to reweight training advantages, and the policy-gradient update remains otherwise unchanged.
At inference time, the model receives only the original prompt without privileged answers.

\begin{algorithm}[h]
\caption{PBSD: Privileged Bayesian Self-Distillation}
\label{alg:pbsd}
\footnotesize
\begin{algorithmic}[1]
\STATE \textbf{Input:} policy $\pi_\theta$, prompts $\mathcal{X}$, privileged answers $\mathcal{R}$
\STATE \textbf{Input:} verifier $V$, group size $G$, soft scale $\eta$, clip range $\gamma$
\STATE \textbf{Output:} reweighted token advantages $\widehat{A}_{j,i}$

\FOR{each update}
    \STATE Sample prompt $x\sim\mathcal{X}$ and roll out
    $\{\tau_j\}_{j=1}^{G}\sim\pi_\theta(\cdot\mid x)$.
    \STATE Compute rewards $s_j\gets V(x,\tau_j)$ and normalized advantages:
    \[
    A_j \gets
    \frac{s_j-\mu_s}{\sigma_s},
    \quad
    \mu_s=\frac{1}{G}\sum_{g=1}^{G}s_g .
    \]

    \FOR{each trajectory $\tau_j$ with privileged answer $r_j$}
        \STATE Form privileged context $x_j^\star$ by augmenting $x$ with $r_j$.
        \STATE Let $z_j$ be the generated token sequence in $\tau_j$.
        \STATE Score generated tokens under two contexts:
        \[
        \begin{aligned}
        \ell^s_{j,i}
        &\gets \log \pi_\theta
        (z_{j,i+1}\mid x,z_{j,\le i}),\\
        \ell^p_{j,\phi(i)}
        &\gets \log \pi_\theta
        (z_{j,i+1}\mid x_j^\star,z_{j,<i+1}),
        \end{aligned}
        \]
        where $\phi(i)$ maps a student loss position to its privileged scoring position.
        \STATE Initialize $\widehat{A}_{j,i}\gets A_j$ for all generated loss tokens.

        \FOR{each assistant turn $k$}
            \STATE Let $\mathcal{T}_{j,k}$ be loss-token positions in turn $k$.
            \STATE Let
            $\mathcal{I}_{j,k}\gets
            \{i\in\mathcal{T}_{j,k}:\phi(i)\text{ exists}\}$.

            \IF{$\mathcal{I}_{j,k}=\emptyset$}
                \STATE $\rho_{j,k}\gets 1$.
            \ELSE
                \STATE Estimate the turn-level privileged likelihood gap:
                \[
                \Delta_{j,k}\gets
                \frac{1}{|\mathcal{I}_{j,k}|}
                \sum_{i\in\mathcal{I}_{j,k}}
                \bigl(\ell^p_{j,\phi(i)}-\ell^s_{j,i}\bigr).
                \]
                \STATE Compute clipped soft correction:
                \[
                c_{j,k}\gets
                \operatorname{clip}_{[-\gamma,\gamma]}
                \left[
                \tanh\left(\frac{\Delta_{j,k}}{\eta}\right)
                \right].
                \]
                \STATE Set turn multiplier:
                \[
                \rho_{j,k}\gets
                1+\operatorname{sgn}(A_j)c_{j,k}.
                \]
            \ENDIF

            \FOR{each $i\in\mathcal{T}_{j,k}$}
                \STATE $\widehat{A}_{j,i}\gets \rho_{j,k}A_j$.
            \ENDFOR
        \ENDFOR
    \ENDFOR

    \STATE Update $\pi_\theta$ with policy-gradient loss using $\widehat{A}_{j,i}$.
\ENDFOR
\end{algorithmic}
\end{algorithm}

\section{Case Study: Turn-Level Credit Assignment}
\label{app:case-study}

We provide two qualitative case studies to illustrate how PBSD assigns
turn-level credit in long-horizon search trajectories. Since the rollout logs
do not store token-level likelihoods, we visualize the privileged likelihood
signal at the assistant-turn level. For each selected turn, we show the model's
reasoning text, the teacher--student likelihood comparison, and a qualitative
credit label. A \emph{High-value} turn is one whose reasoning is aligned with
the answer-conditioned teacher signal, a \emph{Low-value} turn corresponds to
a distracting or misleading reasoning branch, and a \emph{Filtered} turn has a
weak likelihood gap and is therefore not assigned confident positive or
negative credit.

\begin{figure*}[t]
    \centering
    \includegraphics[width=0.95\textwidth]{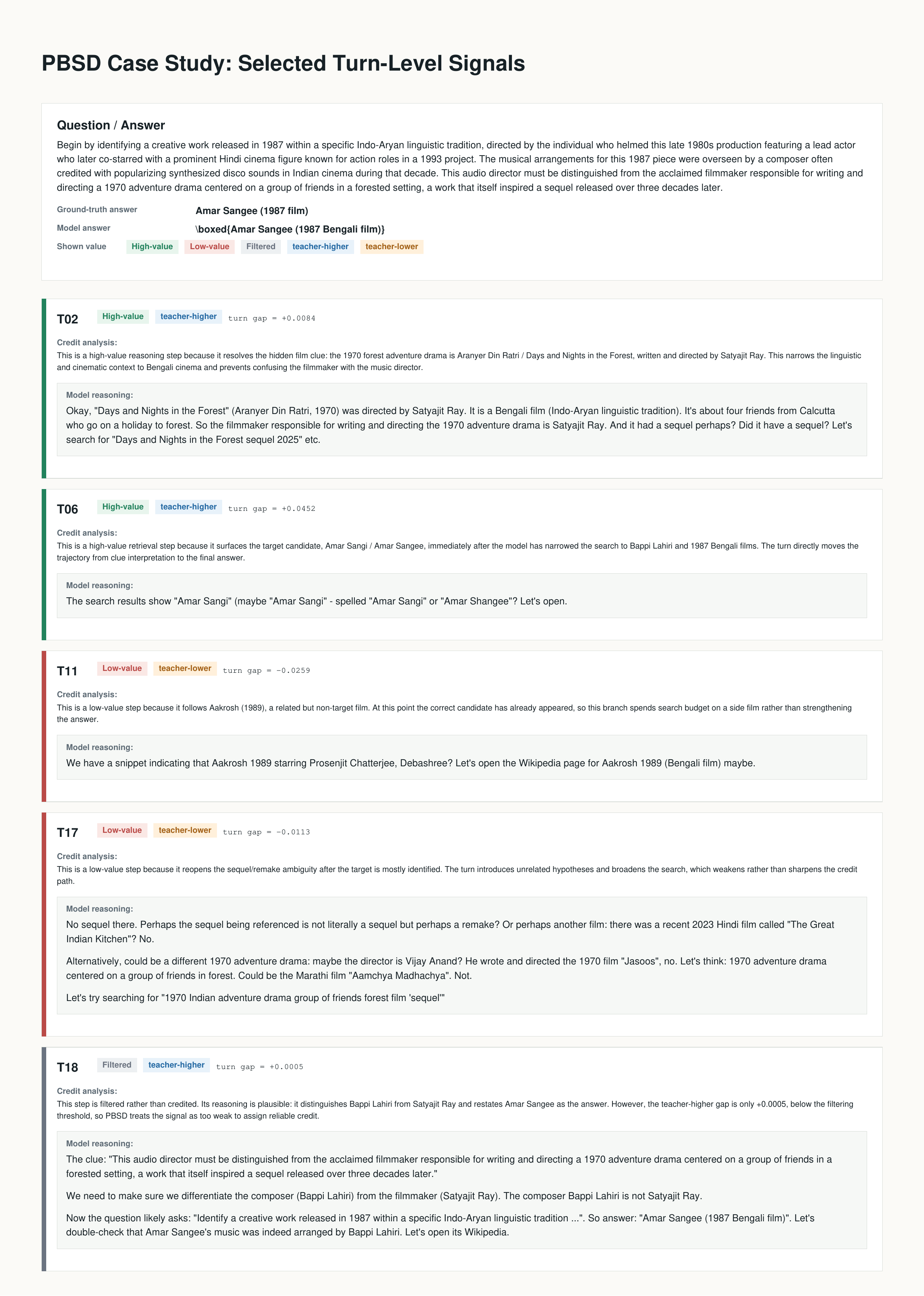}
    \caption{
    Turn-level PBSD signals in a correct rollout. The trajectory reaches the
    correct answer, but PBSD still separates genuinely useful intermediate
    reasoning from redundant or distracting steps.
    }
    \label{fig:pbsd-correct-case}
\end{figure*}

Figure~\ref{fig:pbsd-correct-case} shows a successful trajectory. The model
correctly answers \emph{Amar Sangee (1987 film)}. PBSD identifies turns that
resolve key intermediate clues, such as linking the forest-film clue to
Satyajit Ray and surfacing the candidate title \emph{Amar Sangee}, as
high-value. It also marks redundant or off-path verification steps as
low-value, while filtering weak signals that are plausible but too small to
credit confidently.

\begin{figure*}[t]
    \centering
    \includegraphics[width=0.95\textwidth]{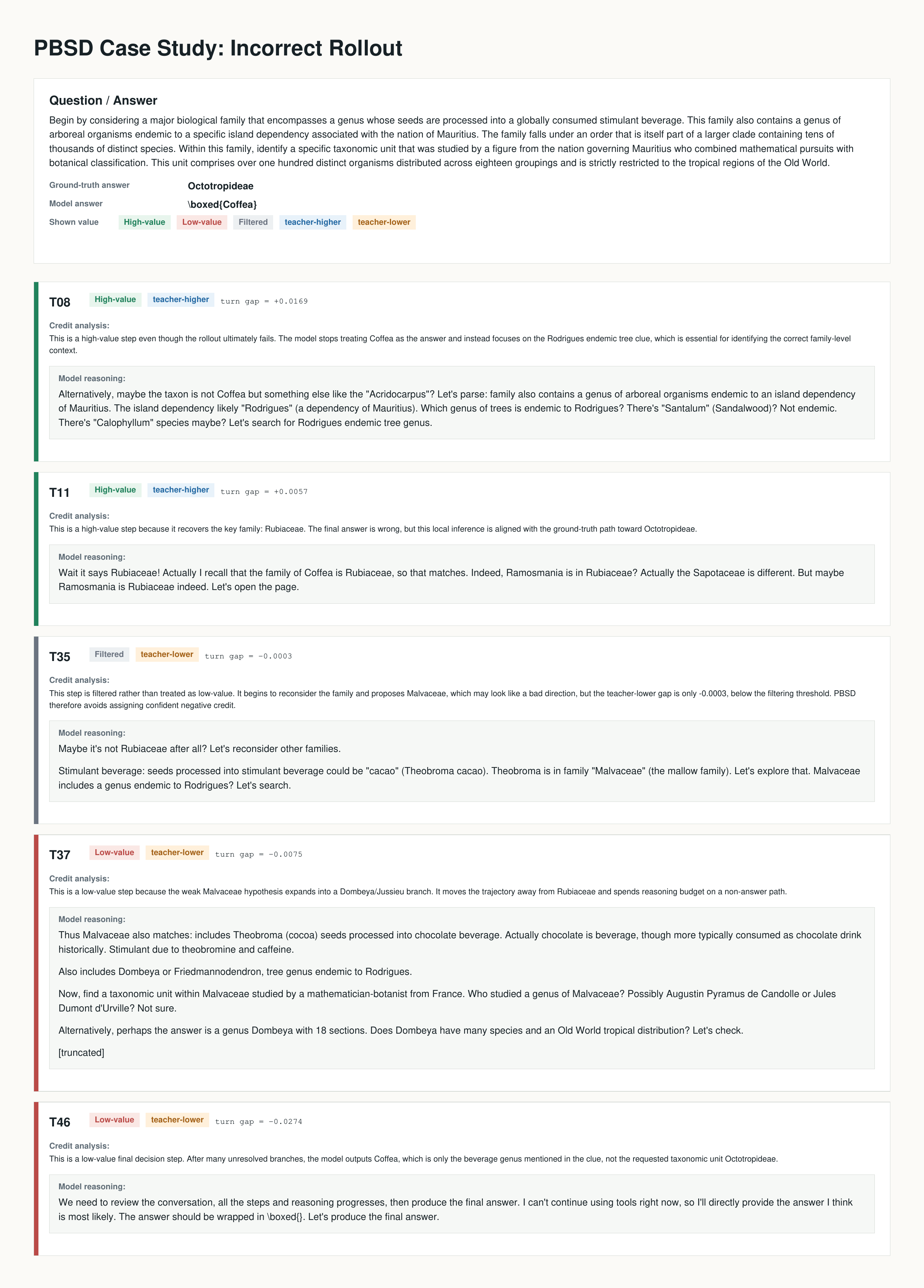}
    \caption{
    Turn-level PBSD signals in an incorrect rollout. Although the final answer
    is wrong, PBSD distinguishes locally useful reasoning steps from misleading
    branches and the final erroneous decision.
    }
    \label{fig:pbsd-incorrect-case}
\end{figure*}

Figure~\ref{fig:pbsd-incorrect-case} shows a failed trajectory. The ground-truth
answer is \emph{Octotropideae}, while the model incorrectly outputs
\emph{Coffea}. Even in this failed rollout, PBSD identifies locally useful
steps, such as recovering the Rubiaceae family and focusing on the Rodrigues
endemic-tree clue. At the same time, it assigns low value to turns that expand
into the wrong Malvaceae/Dombeya branch or collapse to the final incorrect
answer. This illustrates that PBSD does not treat an entire failed trajectory
uniformly: it can preserve useful partial reasoning while penalizing the steps
that drive the model away from the target.

\end{document}